\documentclass[conference]{IEEEtran}
\IEEEoverridecommandlockouts
\usepackage{cite}
\usepackage{amsmath,amssymb,amsfonts}
\usepackage{algorithmic}
\usepackage{graphicx}
\usepackage{textcomp}
\usepackage[dvipsnames, svgnames, x11names]{xcolor}
\usepackage{algorithm}
\usepackage{subfigure}
\usepackage{pgfplots}
\usepackage{multirow}
\usepackage{booktabs}
\usepackage{bm}
\usepackage{makecell}
\usepackage{times}
\usepackage{epsfig}

\def\BibTeX{{\rm B\kern-.05em{\sc i\kern-.025em b}\kern-.08em
    T\kern-.1667em\lower.7ex\hbox{E}\kern-.125emX}}
\begin{document}

\title{SST: Real-time End-to-end Monocular 3D Reconstruction via Sparse Spatial-Temporal Guidance
\thanks{This work was supported by the National Key R\&D Program of China under Grant 2018AAA0102801. Authors with $*$ have equal contributions.}}

\author{
\hspace{-12pt}
Chenyangguang Zhang$^{1,\ast}$, 
Zhiqiang Lou$^{1,\ast}$, 
Yan Di$^{2}$, 
Federico Tombari$^{2, 3}$, and
Xiangyang Ji$^{1}$\\
\\
$^1$ Tsinghua University\quad
$^2$ Technical Univerisity of Munich \quad
$^3$ Google \\
\small{\texttt{\{zcyg22,lzq20\}@mails.tsinghua.edu.cn},\quad \texttt{xyji@tsinghua.edu.cn} } \\
}

\maketitle

\begin{abstract}
Real-time monocular 3D reconstruction is a challenging problem that remains unsolved.
Although recent end-to-end methods demonstrate promising results, tiny structures and geometric boundaries are hardly captured due to their insufficient supervision neglecting spatial details and oversimplified feature fusion ignoring temporal cues.
To address the problems, we propose an end-to-end 3D reconstruction network SST, which utilizes \textit{S}parse estimated points from visual SLAM system as additional \textit{S}patial guidance and fuses \textit{T}emporal features via a cross-modal attention mechanism, achieving more detailed reconstruction results.
We propose a Local Spatial-Temporal Fusion module to exploit more informative spatial-temporal cues from multi-view color information and sparse priors, as well a Global Spatial-Temporal Fusion module to refine the local TSDF volumes with the world-frame model from coarse to fine.
Extensive experiments on ScanNet and 7-Scenes demonstrate that SST outperforms all state-of-the-art competitors, whilst keeping a high inference speed at 59 FPS, enabling real-world applications with real-time requirements.
\end{abstract}

\begin{IEEEkeywords}
3D reconstruction, real time, visual SLAM guidance
\end{IEEEkeywords}

\section{INTRODUCTION}

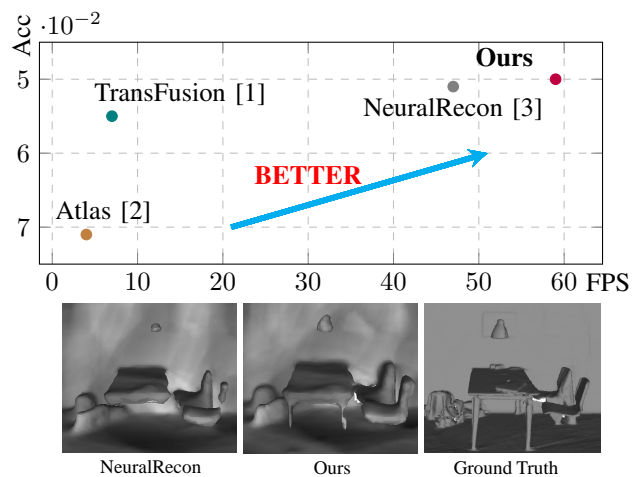
\begin{figure}[t]
    \centering
    \begin{tikzpicture}
    \hspace{-4pt}
    \begin{axis}[xlabel=FPS,ylabel=Acc,
    x label style={at={(axis description cs:1.01,0.18)},anchor=north},
     y label style={at={(axis description cs:0.10,1.03)},anchor=north},
    ymax=0.075, ymin=0.045, legend pos=south east,
    ymajorgrids=true,xmajorgrids=true,
    grid style=dashed,
     y dir=reverse,
    width=0.5\textwidth,
    height=0.25\textwidth]
        \addplot [mark=*, color=purple, ,mark options={fill=purple}]
        coordinates {
    (59, 0.050)
    };
    \addplot [mark=*, color=teal, ,mark options={fill=teal}]
    coordinates {
    (7, 0.055)
    };
    \addplot [mark=*, color=brown, ,mark options={fill=brown}]
    coordinates {
    (4, 0.071)
    };
    \addplot [mark=*, color=gray, ,mark options={fill=gray}]
    coordinates {
    (47, 0.051)
    };
\addplot[thick, cyan,-stealth,quiver={
        u=30, v=-0.01,
        scale arrows=1.0,
        every arrow/.append style={line width=2pt},
    }]   (21, 0.07) ;
    \node at (axis cs:15,0.052){TransFusion~\cite{bozic2021transformerfusion}} ;
    \node at (axis cs:6,0.068) {Atlas~\cite{murez2020atlas}};
    \node at (axis cs:47,0.054) {NeuralRecon~\cite{sun2021neuralrecon}};
    \node at (axis cs:53,0.047) {\textbf{Ours}};
    \node at (axis cs:30,0.063) {\color{Red}\textbf{BETTER}};
    \end{axis}
    \vspace{10pt}
    \end{tikzpicture}
    
    \includegraphics[width=0.40\textwidth]{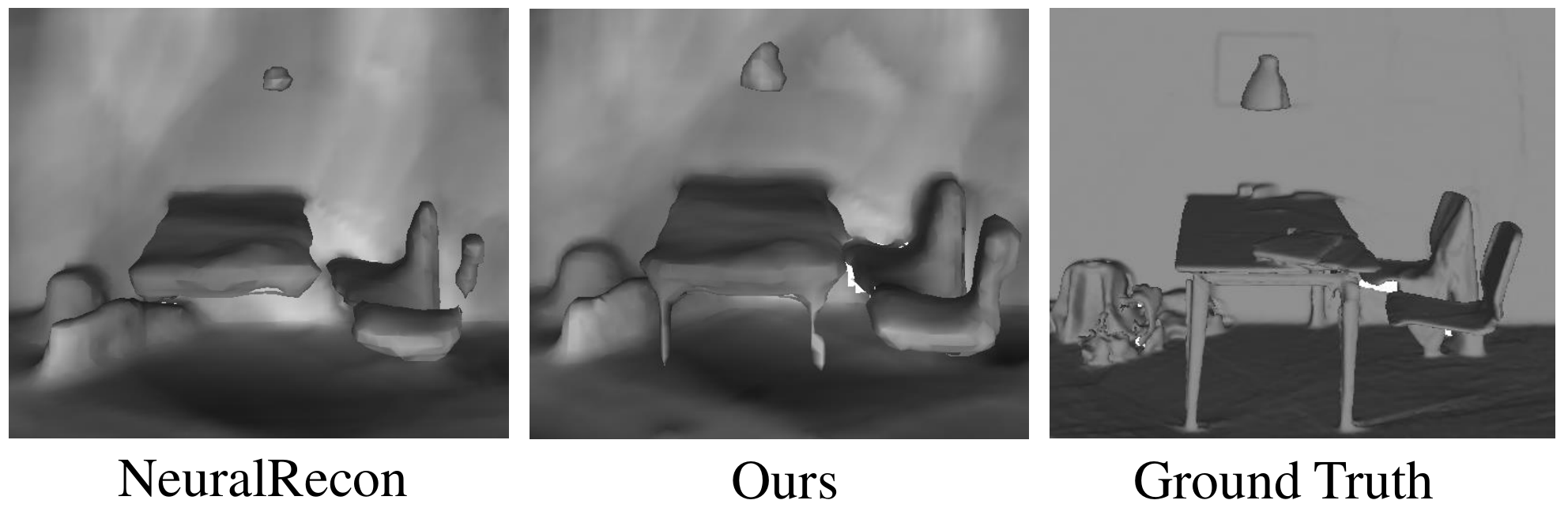}
    \caption{Quantitative and qualitative comparison on ScanNet\cite{dai2017scannet} benchmark. Compared to other competitors, our SST achieves best comprehensive performance, reconstructing more precise tiny structures and geometric boundaries.}
    \label{fig:teaser}
\end{figure}

Monocular 3D scene reconstruction has aroused wide research interest due to its promising applications in AR/VR, robotic manipulation, and scene understanding.
Unlike Lidar or RGB-D methods, monocular 3D reconstruction refers to predicting the 3D scene model from several consecutive frames, without direct distance measurements.
Thus, it is severely ill-posed and remains unsolved yet.

Traditional visual SLAM systems can estimate ego-motion accurately, but can only provide coarse sparse reconstruction results due to high computational consumption and matching ambiguity. 
Recently, deep-learning-based methods have been proposed to achieve satisfactory dense reconstruction performance, which can be divided into two-stage and end-to-end methods. 
Two-stage methods~\cite{duzceker2021deepvideomvs,long2021multi,wang2018mvdepthnet,im2019dpsnet}, occupying the mainstream, first estimate per-frame depth and then fuse the back-projected point clouds.
End-to-end methods directly regress 3D Truncated Signed Distance Function (TSDF) volumes via neural networks.

For end-to-end methods, if well reconstructed geometric details are demanded, intolerable GPU memory cost from high-resolution geometry volumes emerges.
Although sparse volume representation~\cite{sun2021neuralrecon} has been proposed, current methods are still use insufficiently detailed voxel-based representation. 
Thus, they can only be coarsely supervised via ground-truth TSDF volume with large voxel size, losing many of geometry details compared to two-stage methods with dense depth supervision.
To enhance the geometry information neglected by current end-to-end methods, we introduce the sparse depth input, which is a byproduct in ego-motion estimation process of visual SLAM system, and has been proved to contain heuristic geometric guidance~\cite{ma2018sparse,huang2021s3}.
We harness a lightweight CNN to encode the sparse depth input, and then fuse this spatial guidance along with color features.

Besides the drawback of coarse supervision due to voxel scale restriction, current end-to-end methods~\cite{sun2021neuralrecon,murez2020atlas} cannot construct fine 3D feature volumes since they conduct simple average feature fusion, which neglects temporal connections from different observation views.
Thus, considering simultaneously utilizing the sparse cross-modal guidance and the multi-view temporal cues, we carefully design a novel sparse cross-modal attention mechanism.
Integrating sparse spatial-temporal guidance, it is capable to extract more informative 3D feature volumes for accurate reconstruction.

We dedicate ourselves to a real-time monocular 3D reconstruction pipeline SST with the state-of-the-art performance and less computational resource consumption.
The proposed SST follows a economic local-to-global pipeline consisting of three main modules: Feature Extraction (FE), Local Spatial-Temporal Fusion (LSTF) and Global Spatial-Temporal Fusion (GSTF). 
Given a fragment of images, corresponding camera poses and sparse depth measurements from real-time visual SLAM system, SST first extracts pixel-wise cross-modal features from images and sparse depth points respectively, via the FE module. 
Then, after back-projecting features into 3D space, the novel LSTF module distills and aggregates multi-view features locally using learned attention-aware weights to construct effective 3D feature volumes. 
Finally, for globally consistent reconstruction, we design a GSTF module to refine and fuse local TSDF volumes with global 3D model.
We design a well-organized lightweight recurrent unit for GSTF, which plays an indispensable role to accelerate the whole pipeline.
SST follows a coarse-to-fine manner, and the final reconstructed TSDF volumes are outputted at the highest level.

The main contributions are summarized as follows,
\begin{itemize}
\item 
We propose a novel end-to-end network SST for real-time monocular 3D scene reconstruction, which outperforms all real-time competitors and achieves comparable results to the state-of-the-art method but runs 8X faster at 59FPS on ScanNet~\cite{dai2017scannet} and 7-scenes~\cite{shotton2013scene} benchmarks.
\item 
We investigate the shortcomings of previous end-to-end methods and present a novel LSTF module with proposed sparse cross-modal attention mechanism that allows for adaptive feature aggregation for color information and sparse point-based geometry priors, distilling more informative cues for accurate reconstruction.
\item
We design a lightweight recurrent unit for our GSTF module to efficiently fuse local TSDF volumes with global 3D model, reducing time and storage consumption considerably.
\end{itemize}

\section{RELATED WORKS}
\subsection{3D Scene Reconstruction}

There are two streams of methods handing 3D scene reconstruction problem, two-stage and end-to-end methods.
Two-stage methods follow a long-standing reconstruction pipeline that first estimates depth map of each frame and then fuses the back-projected point clouds.
\cite{schonberger2016pixelwise, di2019monocular, di2020unified} are popular traditional methods with a patch matching based pipeline. 
Learning-based approaches \cite{im2019dpsnet,long2021multi,liu2019neural,hou2019multi,duzceker2021deepvideomvs} often build a shared 3D cost volume in the reference camera space using feature averaging.
However, these two-stage methods still confront with spatial inconsistency and redundant computation problems.
Atlas~\cite{murez2020atlas} is the first end-to-end scene reconstruction approach, proposing a global averaging feature volume, but with low efficiency and not incrementally.
TransformerFusion~\cite{bozic2021transformerfusion} and NeuralRecon~\cite{sun2021neuralrecon} are two impressive incremental methods.
\cite{bozic2021transformerfusion} constructs a global feature volume, gaining superior results but suffering from more time and storage consumption.
\cite{sun2021neuralrecon} extracts feature volumes locally and then fuses to the global.
Different from them, our SST fuses multi-frame spatial-temporal features in local coordinate system and then converts them to the global, which is demonstrated to be efficient for 3D reconstruction. 

\subsection{Sparse Geometry Feature Fusion}
Some works~\cite{ma2018sparse,huang2021s3} illustrate noisy sparse points can enhance monocular depth estimation results.
They design upsampling networks to fuse sparse geometry features and RGB features, aiming at producing a completed dense depth map. 
Other related works concentrate on 3D object detection~\cite{yoo20203d,huang2020epnet,xie2020pi, su2023opa} or end-to-end autopilot control task~\cite{prakash2021multi}.
\cite{yoo20203d,huang2020epnet,xie2020pi} focus on unifying representation for multi-modal feature, whether voxel-based or point-based.
We have not found any mature pipeline with respect to fusing sparse geometry information for end-to-end 3D scene reconstruction task.
Our proposed method fills in the gaps of this field and gains impressive results considering both accuracy and speed.

\section{METHOD}

\subsection{Overview}\label{overview}
\begin{figure*}[t]
    \centering
    \includegraphics[width=0.92\textwidth]{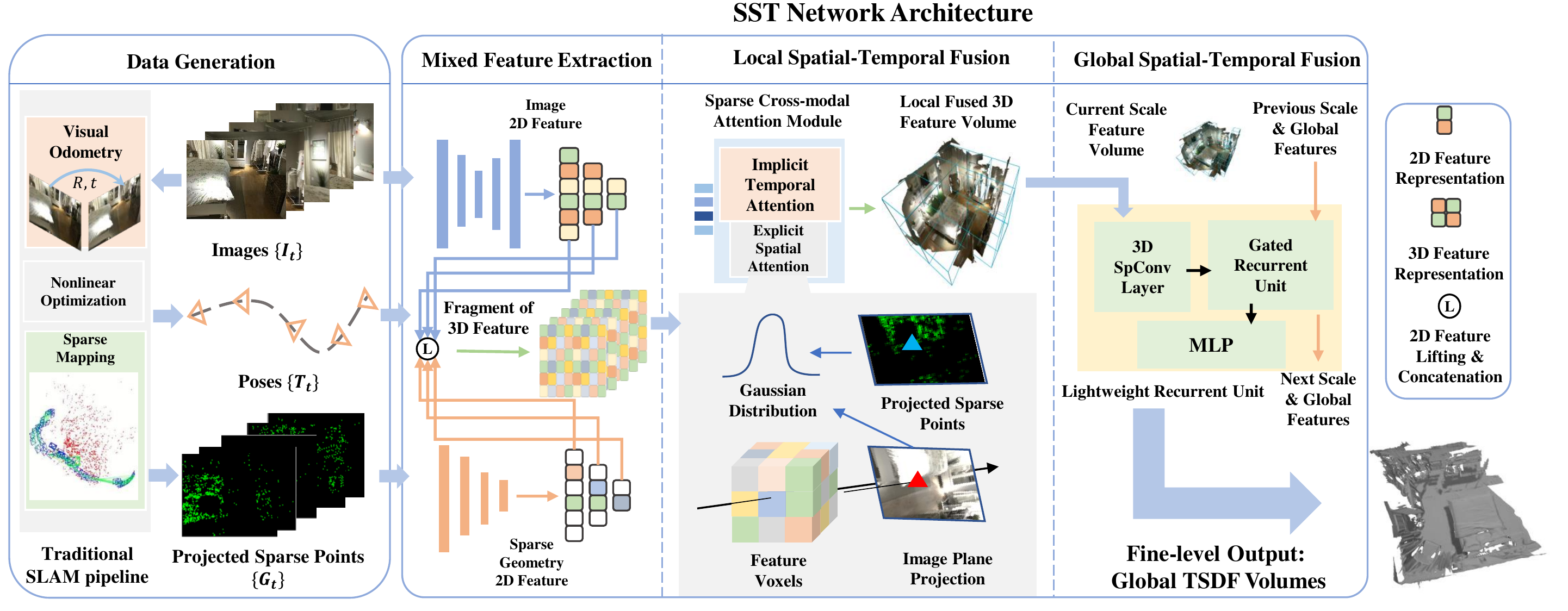}
    \caption{Pipeline Overview of SST.}
    \label{fig:pipeline}
\end{figure*}

As illustrated in Fig.~\ref{fig:pipeline}, given a monocular image sequence $\{I_t\}$, SST first utilizes a visual SLAM system \cite{ORBSLAM2} to generate noisy sparse geometry priors $\{G_t\}$ and corresponding camera poses $\{T_t\}$ as the network inputs.
$\{G_t\}$ consist of projected sparse depth maps $\{SD_t\}$ and corresponding reprojection error maps $\{E_t\}$.
We convert $\{E_t\}$ to confidence maps $\{CO_t\}$ via $CO_t = exp\left(-\lambda E_t\right)$, and then concatenate $\{CO_t\}$ and $\{SD_t\}$ as final geometric priors $\{G_t\}$.

Our SST, consisting of Feature Extraction (FE), Local Spatial-Temporal Fusion (LSTF) and Global Spatial-Temporal Fusion (GSTF) modules, incrementally receives $\{I_t\}$, $\{G_t\}$ and $\{T_t\}$, and reconstructs accurate dense 3D geometry structures in real time.
Specifically, the input sequence will be split into several fragments ${F_l}$ and each fragment contains $N$ frames, predefined as 9 during our implementation.
FE extracts and back-projects features from $I_t$ and $G_t$ for each key frame in the sequence to obtain raw feature volumes $FV_t^{IG}$.
Then LSTF collects features $\{FV_t^{IG}\}_N$ in the current fragment $F_l$ and fuses them into a local feature volume $FV_l^{frag}$ via proposed sparse cross-modal attention mechanism, which strengthens the mixed features by the adaptive weights considering both temporal and spatial guidance.
Finally, $FV_l^{frag}$ will be efficiently updated to the global feature volume $FV^{global}$ in GSTF with the lightweight recurrent unit, followed by simple feed-forward layers to output final TSDF volumes. 
Inspired by \cite{sun2021neuralrecon}, our network repeats the above inference pipeline in a coarse-to-fine manner with three scales and utilizes coarse predictions to guide the high-resolution 3D feature volume for less computation cost.

\subsection{Feature Extraction}
Given a color image $I_t$ and corresponding geometry prior $G_t$, the FE module generates mixed 3D feature volume $FV_t^{IG}$ which contains rich semantic cues from $I_t$ and spatial information from $G_t$ .
According to \cite{merrill2021robust}, we select different feature extraction backbones for $I_t$ and $G_t$ respectively because of their different modalities.
Concretely, a lightweight MnasNet~\cite{tan2019mnasnet} is applied for image feature $F_{t,color}$ extraction. For informative geometry prior input $G_t$, we design a lightweight CNN composed of 4 cascaded convolution blocks to output implicit geometry feature $F_{t,geo}$ with 8 channels.


We then back-project $F_{t,geo}$ and $F_{t,color}$ into 3D space to construct feature volume $FV_t^{IG}$ for key frame $I_t$,  
\begin{equation}
\label{voxel_feature}
FV_t^{IG} = {Backproj}\left({Cat}\left(F_{t,color}, F_{t,geo}\right)\right)
\end{equation}
where $Cat$ means channel-wise concatenation and  $Backproj$ denotes 2D-3D back-projection operation from the image plane to the local frame coordinate.
Note that $FV_t^{IG}$, $F_{t,color}$ and $F_{t,geo}$ are all calculated at 3 resolution scales for our coarse-to-fine inference framework.

\subsection{Local Spatial-Temporal Fusion}\label{LSTF}
 Given $l$-th fragment's 3D features $\{FV_t^{IG}\}_N$ from $N$ consecutive frames $\{I_t\}_N$, the task of LSTF is to output an effective 3D feature volume $FV_l^{frag}$ for current local fragment, similar to local window optimization in traditional SLAM pipelines.

\begin{figure}[ht]
    \centering
    \includegraphics[width=0.48\textwidth]{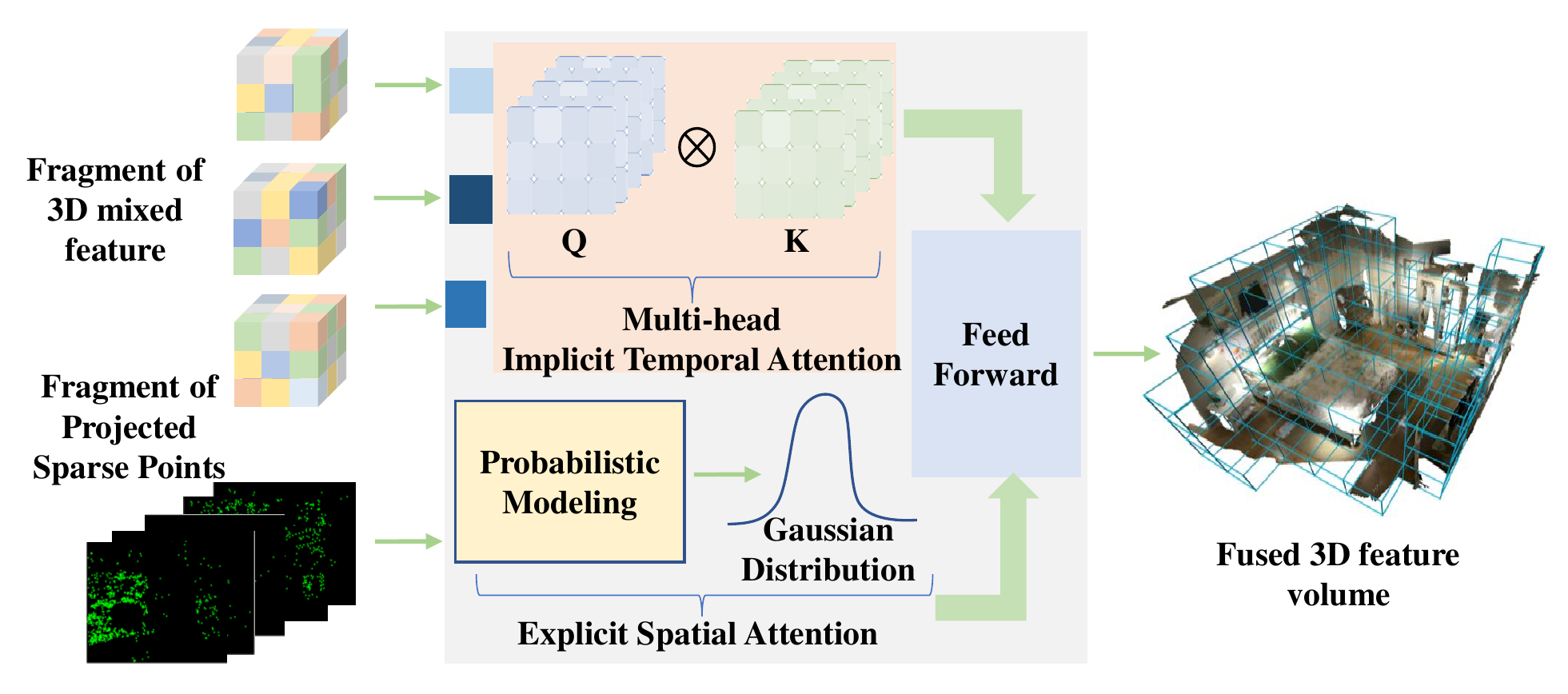}
    \caption{Details of the proposed LSTF.}
    \label{fig:detail_trans}
\end{figure}

For voxel $v_i$ in local frame coordinate system, previous methods~\cite{sun2021neuralrecon,murez2020atlas} directly generate the voxel feature $FV_{l,v_i}^{frag}$ via simple averaging as $\frac{1}{N}\sum_{t=1}^{N}FV_{t,v_i}^{IG}$.
This simple averaging operation loses most spatial information from multi-view features since it cannot learn to attend to important parts for surface reconstruction, i.e. corners, vertical lines, planes etc.
Also, the averaging operation treats each frame in a fragment equally, neglecting temporal correlation among multi-view features.
Furthermore, considering the input geometry priors, if using direct averaging,  $F_{t,geo}$ and $F_{t,color}$ will be processed separately without any interaction, which increases the difficulty of subsequent module to learn informative fusion features for accurate reconstruction.
To handle these shortcomings, as shown in Fig. \ref{fig:detail_trans}, we propose LSTF that utilizes the proposed sparse cross-modal attention mechanism, enabling both adaptive weighted feature fusion in temporal dimension and channel-wise multi-modal feature interaction for spatial feature fusion.
The inference procedure is shown as,
\begin{equation}
    \label{eq:attn}
    \begin{aligned}
    &A_{in} = \left[FV_{1,v_i}^{IG},\cdots,FV_{N,v_i}^{IG}\right]\\
    &Q = W_q A_{in},\  K=W_k A_{in},\  V=W_v A_{in} \\
    &\omega_{l,im}  = Softmax(Q K^T),\ A_{out} = \omega_{l,im} \omega_{l,ex} V \\
    \end{aligned}
\end{equation}
where $A_{in},A_{out}, W_* $ denote the input, output, and learnable parameters of the attention module.
The final attention of the sparse cross-modal attention mechanism is the mixture of implicit temporal attention $\omega_{l,im}$ and explicit spatial attention $\omega_{l,ex}$.
To compute $\omega_{l,ex}$, for the weight $\omega_{l,i,t}$ of voxel $v_i$ in the $t$-th view of this fragment, we first project $v_i$ on the image plane of the $t$-th view to get the projected depth $d_{v_i}$ and pixel position $p_{v_i}$, and $\omega_{l,i,t}$ is computed as
\begin{equation}
    \begin{aligned}
    \omega_{l,i,t} & = \left\{
    \begin{matrix}
    Gauss_{\sigma_{E_{t}p_{(v_i)}}}(\Vert  SD_{t}(p_{v_i})-d_{v_i}\Vert) &  SD_{t}(p_{v_i})>=0\\
    1 & otherwise \\
    \end{matrix}
    \right.
    \end{aligned}
\end{equation}
where $SD_{t}(p_{v_i})$ means corresponding sparse depth value of $p_{v_i}$ in $SD_{t}$. $\omega_{l,ex}$ tends to assign higher weight for voxels near sparse points and the learned adaptive weight $\omega_{l,im}$ can attenuate the influence of unnecessary features and persevere important cues, together assisting in constructing effective feature volumes.
The value weight matrix $W_v$ further mixes features channels from geometry priors and color images, integrating sparse spatial information into final output features, conducting spatial fusion process.
The fragment feature $FV_{l,v_i}^{frag}$ of voxel $v_i$ is computed as $FV_{l,v_i}^{frag} = \mathbf{LSTF} (FV_{1,v_i}^{IG},\cdots,FV_{N,v_i}^{IG})$.

\subsection{Global Spatial-Temporal Fusion}\label{GSTF}

\begin{figure}[ht]
    \centering
    \includegraphics[width=0.48\textwidth]{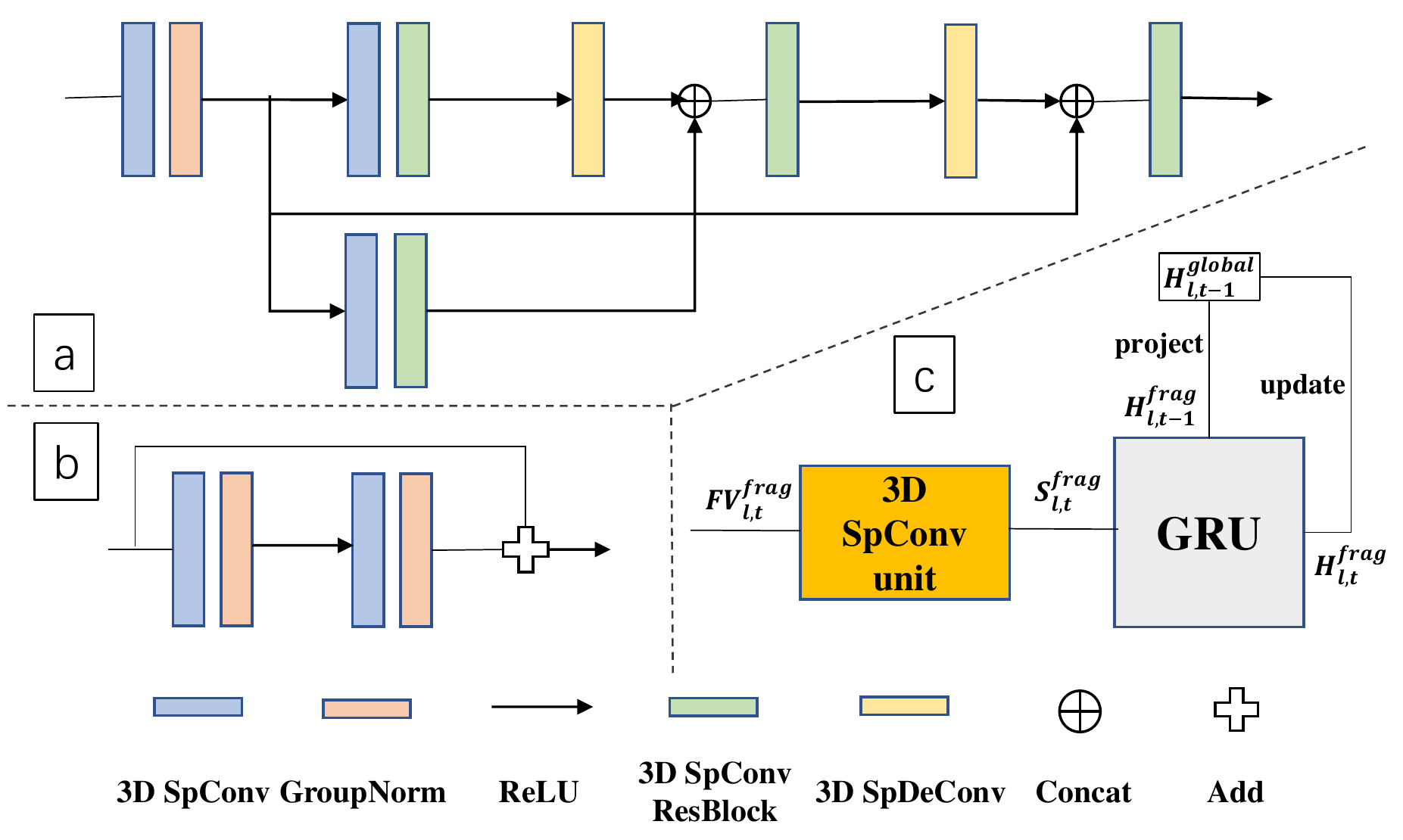}
    \caption{Details of the GSTF module.}
    \label{fig:GSTF}
\end{figure}

To enable consistent incremental reconstruction, we adopt a proposed lightweight recurrent unit for fusing current fragment feature volume $FV_{l}^{frag}$ into previous established global feature volume, conceptually similar to global optimization in traditional SLAM pipelines.
Specifically, the lightweight recurrent unit is a 3D sparse convolutional variant of Gated Recurrent Unit (GRU). 
It maintains a global hidden state volume $H_{l, t-1}^{global}$, and updates it as follows.
First, $FV_{l, t}^{frag}$ is input to 3D sparse convolution layers for extracting its corresponding surface geometry feature $S_{l, t}^{frag}$.
Then hidden state volume $H_{l, t-1}^{frag}$ in local coordinate system is generated from $H_{l, t-1}^{global}$ and fused with $S_{l, t}^{frag}$, resulting in updated local hidden state volume $H_{l, t}^{frag}$ then transformed into  the global volume $H_{l, t}^{global}$.
Two simple MLP layers are applied to $H_{l, t}^{global}$ to estimate occupancy grid $O_l^{global}$ and TSDF volume $TSDF_l^{global}$ respectively. The whole pipeline is shown in Fig.~\ref{fig:GSTF} (c).

During implementation, we find the inference time bottleneck lies in the structure of 3D sparse convolution layers. Thus, we design a new lightweight structure with much higher inference speed compared to \cite{sun2021neuralrecon}, in which the 3D sparse convolution layers are redundant and less organized. We refer to the implementation adopted by \cite{tang2020searching} but significantly modify the network structure and reduce the number of parameters. The concrete structure of the 3D sparse convolution (3D SpConv) unit design is exhibited in Fig.~\ref{fig:GSTF} (a). Fig.~\ref{fig:GSTF} (b) shows the detail of the 3D SpConv residual block, which is simplified by us compared with \cite{tang2020searching}. In practice, the novel lightweight recurrent unit maintains the quality of reconstruction but distinctively reduces inference time, playing a pivotal role in accelerating SST network. 

\section{EXPERIMENTS}
\begin{figure*}[tbp]
    \centering
    \includegraphics[width=0.85\textwidth]{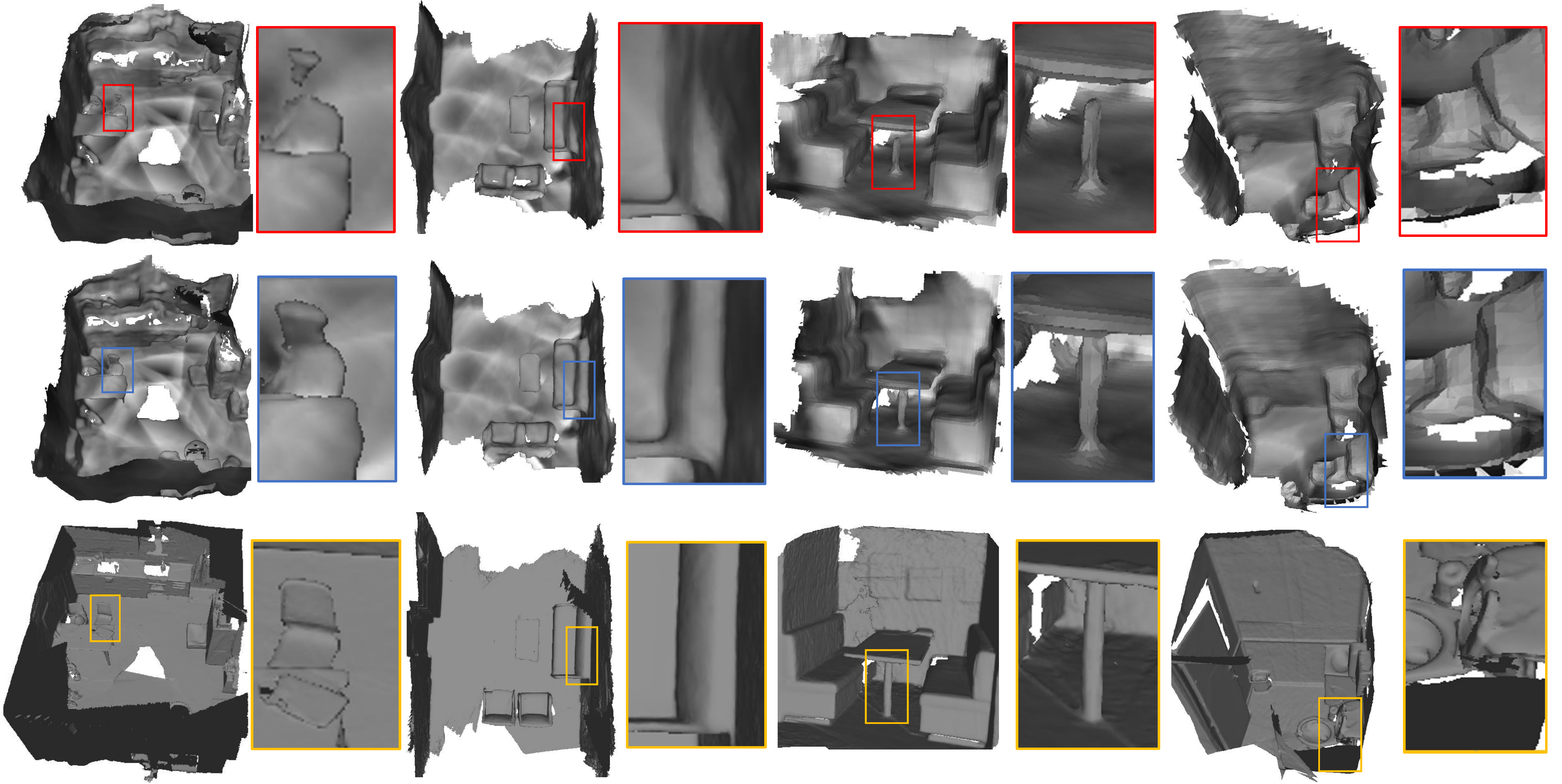}
    \caption{Qualitative comparison on ScanNet. From top to down are the results of NeuralRecon\cite{sun2021neuralrecon}, our results and ground truth respectively. Compared to real-time state-of-the-art \cite{sun2021neuralrecon}, our SST presents more accurate and coherent reconstruction results.}
    \label{fig:exp_results}
\end{figure*}

\subsection{Experimental Setup}
\textbf{Datasets and Baselines.}
The experiments are performed on two common benchmarks, ScanNet~\cite{dai2017scannet} and 7-Scenes~\cite{shotton2013scene}. 
We use the same training and validation splits with previous methods~\cite{murez2020atlas}\cite{sun2021neuralrecon}\cite{bozic2021transformerfusion}.
Following~\cite{long2020occlusion}\cite{sun2021neuralrecon}, we directly apply our model trained on ScanNet for evaluation on 7-Scenes to validate the generalization ability.
We compare our method with two-stage MVS methods and end-to-end reconstruction methods. 
The two-stage methods contain MVDNet~\cite{wang2018mvdepthnet}, GPMVS~\cite{hou2019multi}, DPSNet~\cite{im2019dpsnet} and COLMAP~\cite{schonberger2016pixelwise}, 
while the end-to-end methods include non-real-time Atlas~\cite{murez2020atlas}, TransFusion~\cite{bozic2021transformerfusion} and real-time NeuralRecon~\cite{sun2021neuralrecon}.

\textbf{Metrics.}
Current real-time state-of-the-art method~\cite{sun2021neuralrecon} and its non-real-time counterpart~\cite{bozic2021transformerfusion} follow different 3D metrics to evaluate their methods on ScanNet. 
Thereby, we need to make comparison to them under two corresponding 3D metrics respectively. 
The 3D metrics exhibited in the top block of Tab. \ref{tab:3d_scannet} follow \cite{sun2021neuralrecon}. 
In the bottom block of Tab. \ref{tab:3d_scannet}, we present metrics following \cite{bozic2021transformerfusion}.
For evaluation under 2D depth metrics, we render the reconstructed mesh to the image plane and obtain depth estimations.
All 2D depth metrics for ScanNet evaluation in Tab. \ref{tab:2d_scannet} are defined in \cite{sun2021neuralrecon}\cite{eigen2014depth}. 
For metrics in the generalization validation on 7-Scenes, we also follow metrics in \cite{sun2021neuralrecon} as presented in Tab. \ref{tab:7_scenes}.

\textbf{Implementation Details.}
In the FE module, the MnasNet extracts color feature with 80, 40 and 24 channels for three coarse-to-fine scales.
We initialize the MnasNet via weights pretrained from ImageNet~\cite{deng2009imagenet}.
In the LSTF module, after the calculation for sparse cross-modal attention, the used feed-forward layer remains the channels of output local fragment feature volume with 80, 40 and 24.
The lightweight 3D sparse convolution utilized in GSTF is implemented by torchsparse~\cite{tang2020searching} due to its best synthetic performance.
For sparse depth input, we ignore pixels with depth deeper than 3.0m due to their large noise. 
For the voxel output, we predict the occupancy via a Sigmoid layer and the TSDF value via a full-connection layer from the global feature volume.
As in previous end-to-end methods~\cite{sun2021neuralrecon, bozic2021transformerfusion,murez2020atlas}, the voxel size of the finest level is 4cm and TSDF truncation distance $\lambda$ is set to 12cm. 
We adopt the binary cross-entropy loss for the predicted occupancy score and \emph{$l_1$} loss for the log-transform results of regressed TSDF results.
Both loss functions are applied in the three coarse-to-fine levels.

\subsection{Evaluation Results}
\begin{table}[]
\caption{3D geometry metrics on ScanNet.}
\setlength\tabcolsep{4pt}
\begin{center}
\begin{tabular}{ccccccc}
\hline
Method            & Comp          & Acc           & Recall        & Prec          & F-score       &FPS\\
\hline
MVDNet\cite{wang2018mvdepthnet}        & \underline{0.040}          & 0.240          & \underline{0.831}          & 0.208          & 0.329  &28         \\
GPMVS\cite{hou2019multi}             & \textbf{0.031} & 0.879          & \textbf{0.871} & 0.188          & 0.304  &27    \\
DPSNet\cite{im2019dpsnet}            & 0.045          & 0.284          & 0.793          & 0.223          & 0.344    &4  \\
COLMAP\cite{schonberger2016pixelwise}            & 0.069          & 0.135          & 0.634          & 0.505          & 0.558    &0.4 \\
NeuralRecon\cite{sun2021neuralrecon}       & 0.138          & \underline{0.053}          & 0.472          & \underline{0.687}          & \underline{0.559}  &\underline{47} \\
Ours              & 0.124          & \textbf{0.053} & 0.505          & \textbf{0.695} & \textbf{0.584} &\textbf{59}\\
\hline
TransFusion\cite{bozic2021transformerfusion} & 0.082          & 0.055          & \underline{0.600}          & \textbf{0.728}          & \textbf{0.655}  & 7       \\
Atlas\cite{murez2020atlas}             & 0.076          & 0.071          & \textbf{0.605}          & 0.675          & 0.636  & 4        \\
NeuralRecon\cite{sun2021neuralrecon}       &\underline{0.075}           &\underline{0.051}                &0.556    &0.706              &0.621    &\underline{47}      \\
Ours    &\textbf{0.071}&    \textbf{0.050}&        0.584&      \underline{0.714}&     \underline{0.643}           &   \textbf{59}\\
\hline
\end{tabular}
\end{center}
\label{tab:3d_scannet}
\end{table}

\begin{table}[]
\caption{2D depth metrics on ScanNet.} 
\setlength\tabcolsep{4pt}
\begin{center}
\begin{tabular}{cccccc}
\hline
Method  & Abs.Rel.       & Abs.Diff.      & Sq.Rel.        & RMSE          & $\delta<1.25$ \\
\hline
MVDNet\cite{wang2018mvdepthnet}& 0.098          & 0.191          & 0.061          & 0.293          & 89.6\\
GPMVS\cite{hou2019multi}   & 0.130          & 0.239          & 0.339          & 0.472          & 90.6\\
DPSNet\cite{im2019dpsnet}  & 0.087          & 0.158          & 0.035          & 0.232          & 92.5\\
COLMAP\cite{schonberger2016pixelwise}  & 0.137          & 0.264          & 0.138          & 0.502          & 83.4\\
Atlas\cite{murez2020atlas}   & 0.065          & 0.123          & 0.045          & 0.251          & 93.6\\
NeuralRecon\cite{sun2021neuralrecon}& 0.065          & 0.099          & 0.034          & 0.197          & 93.7\\
Ours        & \textbf{0.060} & \textbf{0.092} & \textbf{0.034} & \textbf{0.185} & \textbf{94.0}\\
\hline
\end{tabular}
\end{center}
\label{tab:2d_scannet}
\end{table}

\textbf{ScanNet.}
2D depth metrics and 3D geometry metrics are evaluated on the ScanNet\cite{dai2017scannet} dataset. Tab. \ref{tab:3d_scannet} exhibits quantitative comparison under two categories of 3D geometry metrics followed \cite{sun2021neuralrecon} and \cite{bozic2021transformerfusion} respectively. 

For 3D metrics consistent with \cite{sun2021neuralrecon}, our SST yields best Accuracy, Precision and F-score.
Compared with recent state-of-the-art real-time method~\cite{sun2021neuralrecon}, SST surpasses in all 3D metrics, exceeding \cite{sun2021neuralrecon} in F-score, Precision and Recall by $2.5\%$, $0.8\%$ and $3.3\%$ respectively.
Compared with two-stage methods, SST surpasses the best two-stage method~\cite{schonberger2016pixelwise} in Accuracy by nearly $61\%$, with the promotion of Precision by $19.0\%$.
For 3D metrics consistent with \cite{bozic2021transformerfusion}, SST yields best Completeness and Accuracy as well as second best Precision and F-score.
We surpass the state-of-the-art real-time method \cite{sun2021neuralrecon} (47 FPS) in all 3D metrics. 
SST exceeds \cite{sun2021neuralrecon} in F-score, Precision and Recall by $2.2\%$, $0.8\%$ and $2.8\%$ respectively.
Compared with non-real-time method \cite{murez2020atlas} (4 FPS), we get a slightly inferior Recall since it is an off-line method with advantage of having a global context to complete their previous TSDF predictions. 
For Precision and F-score, SST surpasses \cite{murez2020atlas} by $3.9\%$ and $0.7\%$ respectively.
Compared with state-of-the-art non-real-time method~\cite{bozic2021transformerfusion} (7 FPS), we surpass it in Completeness and Accuracy by $13.4\%$ and $9.1\%$, while get slightly inferior Precision and Recall by only $1.6\%$ and $1.4\%$. 
However, \cite{bozic2021transformerfusion} depends on many cascaded transformer blocks, resulting in great storage and non-real-time performance (7 FPS), while our SST maintains lightweight architecture and real-time inference performance (59 FPS).

For 2D depth metrics shown in Tab. \ref{tab:2d_scannet}, SST outperforms all end-to-end and two-stage competitors.
Compared with the state-of-the-art real-time end-to-end method \cite{sun2021neuralrecon}, SST exceeds in Abs. Rel. by $7.7\%$, Abs. Diff. by $7.1\%$ and RMSE by $6.1\%$.
When compared with two-stage methods, our method also has significantly better performance.

We also compare visualization results in Fig. \ref{fig:exp_results}, illustrating our method can reconstruct more accurate and consistent surface geometry than previous state-of-the-art real-time method NeuralRecon~\cite{sun2021neuralrecon}, especially at regions of tiny structures, e.g. the unbroken table leg and the vertical back of chair.

\textbf{7-Scenes.}
To demonstrate the generalization ability of SST, we utilize 7-Scenes dataset to evaluate our model trained on ScanNet.
Our method still achieves outperforming performance to the state-of-the-art real-time method~\cite{sun2021neuralrecon} (Tab. \ref{tab:7_scenes}). 

\begin{table}[]
\caption{3D metrics on 7-Scenes.}
\setlength\tabcolsep{4pt}
\begin{center}
\begin{tabular}{ccccccc}
\hline
Method            & Comp          & Acc           & Recall        & Prec          & F-score       &FPS\\
\hline
NeuralRecon\cite{sun2021neuralrecon}       & 0.228          & \textbf{0.100}          & 0.228          & 0.389          & 0.282          &47\\
Ours        &\textbf{0.225}          &  0.104    & \textbf{0.242}   &  \textbf{0.392} & \textbf{0.298} &\textbf{59}\\
\hline
\end{tabular}
\end{center}
\label{tab:7_scenes}
\end{table}

\textbf{Efficiency.}
In Tab. \ref{tab:3d_scannet}, we report average running frames-per-second (FPS) measured on an NVIDIA RTX 3090 GPU for all methods.
Our method achieves the highest inference speed of 59 FPS among all competitors (8X faster than the state-of-the-art method \cite{bozic2021transformerfusion}), enabling applications with real-time requirement, e.g. AR on mobile devices.

\subsection{Ablation Studies}\label{ablation}
\setlength\tabcolsep{4pt}
\begin{table}[]
\caption{LSTF ,FE and GSTF architecture ablations on ScanNet under 3D metrics along with the top block of Tab. \ref{tab:3d_scannet}.}
\begin{center}
\begin{tabular}{cccc|cccc}
\hline
  & SCAM  & \thead{Geometry \\Priors} & LWRU & Recall & Prec & F-score & FPS\\
 \hline
a & ×  & ×   & ×   & 0.472   & 0.687 & 0.559 & 47   \\
b & \checkmark   & ×   & ×    & 0.494   & 0.690 & 0.574 & 41   \\
c & \checkmark   & ×   & \checkmark    & 0.495   & 0.695 & 0.576  & \textbf{62}   \\
d & \checkmark   & \checkmark & ×    & 0.496   & 0.695 & 0.579 & 38  \\
e & \checkmark   & \checkmark & \checkmark   & \textbf{0.505}   & \textbf{0.695} & \textbf{0.584} & 59  \\
\hline
\end{tabular}
\end{center}
\label{tab:ablation}
\end{table}

In this section, we conduct ablation experiments on ScanNet to demonstrate the effectiveness of our proposed architecture for making full use of spatial-temporal information for 3D reconstruction.
The proposed sparse cross-modal attention mechanism (SCAM) in LSTF, sparse geometry priors encoder in FE and the lightweight recurrent unit (LWRU) in GSTF are three main concerns of ablations we conduct. 
Results shown in Tab. \ref{tab:ablation} demonstrate significant effectiveness of three modules above. 

Comparing (a) (b), our pipeline with the proposed SCAM yields higher 3D geometry metrics, promoting F-score by $1.7\%$. 
The geometry priors encoder plays another important role in exploiting the sparse spatial guidance.
Concretely, comparing (c) (e), we find that encoding the sparse geometry priors by our proposed CNN promotes the F-score by $1.0\%$.
The proposed LWRU plays a pivotal role in inference speed enhancement of SST, as is illustrated in (d) and (e). 
With the design of our LWRU, the promotion of FPS reaches near $50\%$, and the F-score is slightly enhanced simultaneously.

\section{CONCLUSIONS}
We introduce a real-time monocular 3D reconstruction network SST, yielding accurate reconstruction results whilst keeping high inference speed.
The key idea is to fully utilize spatial-temporal guidance provided by multi-view features and sparse geometry priors.
We design a LSTF module to treat multi-view feature attentively, simultaneously fusing sparse spatial information, and a efficient GSTF module to maintain global consistency of reconstruction.
Extensive experiments demonstrate that our SST network achieves comparable results to the state-of-the-art method but runs 8X faster.
We believe that our local-to-global method with sparse spatial-temporal guidance can inspire future researches aiming at end-to-end 3D reconstruction task.

\bibliographystyle{IEEEtran}
\bibliography{new}

\end{document}